\documentclass[10pt,twocolumn,letterpaper]{article}

\usepackage{wacv}
\usepackage{times}
\usepackage{epsfig}
\usepackage{graphicx}
\usepackage{amsmath}
\usepackage{amssymb}
\usepackage{float}
\usepackage{fixltx2e}
 
\DeclareMathOperator*{\argmin}{arg\,min}
\graphicspath{ {C:\Users\annadani\Desktop} }

\wacvfinalcopy 

\def\wacvPaperID{****} 

\ifwacvfinal\fi
\begin{document}

\title{Sliding Dictionary Based Sparse Representation For Action Recognition}

\author{Yashas Annadani \\
\small National Institute of Technology-Karnataka\\
{\tt\small yashas\_13ee152@nitk.edu.in}
\and
Rakshith D L \\
\small University of Southern California\\
{\tt\small rakshithdl@gmail.com}
\and
Soma Biswas\\
\small Indian Institute of Science, Bangalore\\
{\tt\small soma.biswas@ee.iisc.ernet.in}
}

\maketitle
\ifwacvfinal\thispagestyle{empty}\fi

\begin{abstract}
The task of action recognition has been in the forefront of research, given its applications in gaming, surveillance and health care. 
In this work, we propose a simple, yet very effective approach which works seamlessly for both offline and online action recognition using the skeletal joints.
We construct a sliding dictionary which has the training data along with their time stamps.
This is used to compute the sparse coefficients of the input action sequence which is divided into overlapping windows and each window gives a probability score for each action class.
In addition, we compute another simple feature, which calibrates each of the action sequences to the training sequences, and models the deviation of the action from the each of the training data.
Finally, a score level fusion of the two heterogeneous but complementary features for each window is obtained and the scores for the available windows are successively combined to give the confidence scores of each action class.
This way of combining the scores makes the approach suitable for scenarios where only part of the sequence is available.
Extensive experimental evaluation on three publicly available datasets shows the effectiveness of the proposed approach for both offline and online action recognition tasks.
\end{abstract}


\section{Introduction}

\begin{figure}[t]
	\centering
	\includegraphics[width=1.0\linewidth]{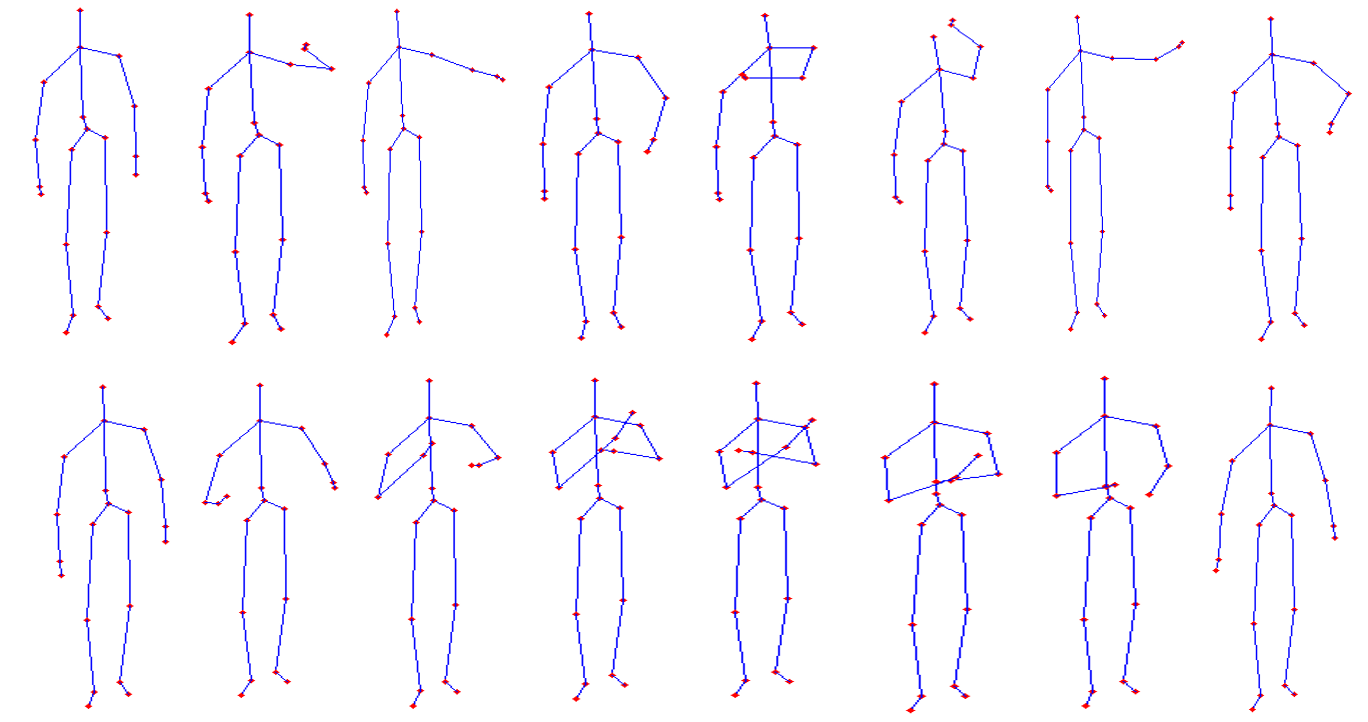}
	\caption{Sample skeletal joints (the red dots joined by blue lines for visualization) for two actions, draw circle (top) and cross arms in the chest (bottom) from the UTD dataset~\cite{utd}.}
\label{intro}
\end{figure}

There has been considerable interest in the computer vision community to recognize human actions from 3D data ever since the inception of Kinect sensor. Kinect provides  multi-channel data which has paved unprecedented avenues to problems of action recognition in particular, and computer vision in general. 
This interest has grown manifold after the work of Shotton {\it et al.}~\cite{shotton2013real}, which estimates the 3D joint locations of humans in real time from a single depth image. 
Although the joint positions obtained from Kinect can be noisy, especially in presence of partial occlusions, the relative simplicity and compact representation offered by 3D skeletal joints have prompted researchers to exploit the advantages arising from the compactness. 
Since then, there has been an increase in research on action recognition by modeling 3D skeletal joints and recent advances~\cite{hussein2013human}\cite{vemulapalli2014human} have indicated that 3D skeletal joints (Figure~\ref{intro}) are a better, simple yet efficient way of representing human actions.   
Online action recognition, where we may be required to recognize the action from partial data also finds wide applications in user interface and in gaming~\cite{kviatkovsky2014online}.

Here, we propose a simple, yet very effective approach which works seamlessly for both offline and online action recognition using the 3D skeletal joints.
The proposed approach is based on learning sparse representations based on a sliding dictionary which is constructed from the training data utilizing their time stamps.
Given an input action sequence, it is divided into several overlapping windows and for each window, we compute its probability of belonging to the different action classes.
The reconstruction error for each class computed with the sparse coefficients is used to compute this probability.
In addition, we compute another simple feature, which calibrates each of the action sequences to the training sequences, and models the deviation of the action from the each of the training data.
The combined score for each window is computed using a score level fusion of these two features and the scores for all the available windows are successively combined to give the probability of each action class.
Extensive experimental evaluation on three publicly available datasets, namely UTD,  UT Kinect and MSRC 12 datasets and comparisons with the state-of-the-art shows the effectiveness of the proposed approach for both offline and online action recognition tasks.
The main contributions of the proposed work are as follows:
\begin{enumerate}
\item Proposed a sliding dictionary based sparse representation framework for action recognition.
\item Approach can work seamlessly for both offline and online action recognition.
\item Simple, yet very effective approach as justified by results on three datasets.
\end{enumerate}

The rest of the paper is organized as follows. 
Section~\ref{related} gives a brief review of the related work.
The proposed algorithm in described in Section~\ref{prop}. 
The experimental results are reported in Section~\ref{expts} and the paper concludes with discussion and future work.


\section{Related Work}
\label{related}
A diverse set of approaches to action recognition using 3D skeletal joints exists in literature. 
Yang and Tian~\cite{yang2012eigenjoints} uses a Naive-Bayes Nearest Neighbour classifier on an offset generated by taking the joint positions difference between specific frames. 
The feature space is scaled down to a lower dimension using PCA. 
Xia {\em et al.}~\cite{xia2012view} model postures by projecting the 3D joints to 3D bins of a histogram (HOJ3D) and  Hidden markov model is used for classification. 
\cite{wang2012mining} employ a multiple kernel learning method to extract the most informative of 3D joint pairs and each of these joint pairs is modeled according to the relative positions with respect to each other. 
Vemulapalli {\em et al.}~\cite{vemulapalli2014human} use dynamic time warping to account for rate variations. 
They model human actions as curves in a lie group and Fourier Temporal Pyramid is employed to handle the temporal misalignment, after warping the curves obtained for each class to its nominal curve. 
One-versus-all linear SVM is used for classification. 
This method gives good results, but the numerical and computational complexities arising from this way of modeling makes it less feasible. 
\cite{hussein2013human} built covaraince of 3D joints (Cov3DJ) and then compute Cov3DJ on different temporal hierarchy to account for order of motion in time. 

Zhu \etal~\cite{zhu2013fusing} perform a feature level fusion of spatiotemporal features and 3D joints features using random forests. 
Bloom \etal~\cite{bloom2012g3d} extract multiple types of features from 3D joints which can be computed in real time: pairwise joint difference, joint velocities with respect to different frames and their magnitudes, and joint angles between three joints. These features obtained from all the joints are concatenated to get a single feature vector.  We refer the reader to \cite{aggarwal2014human} for a detailed discussion on action recognition using 3D skeletal joints. 
Kviatkovsky \etal~\cite{kviatkovsky2014online} uses the covariance descriptor for action recognition by extending it to spatio-temporal domain. 
It is extended to online action recognition by creating a buffer of features extracted using the covariance descriptor.
The buffer is updated when a new frame is added and on demand nearest neighbour is used for classsification.
A heirarchy of bio-inspired multiple skeletal configurations is used in~\cite{chaudhry2013bio} such that each of the configurations represents the motion of set of joints at a particular temporal scale. 
These skeletal configurations are modeled as Linear Dynamic Systems.
Li \etal~\cite{li2010action} build an action graph of sampled 3D points from depth maps, such that each node in an action graph represents a posture common to set of actions to be classified.
Gaussian Mixture Model is used to model the distribution of the sampled 3D points.
In~\cite{ofli2014sequence}, the most informative 3D joints in an action sequence is extracted, where the information is a measure of variance of joint angles in time series. 
Action is represented as a sequence of these sampled informative 3D points and SVM is used for classification.
Ye \etal~\cite{ye2013survey} overviews different action recognition approaches related to skeletal representation and depth maps, and compares the performance of different algorithms on various standard datasets in literature.

\begin{figure*}[t]
	\centering
	\includegraphics[width=1.0\linewidth]{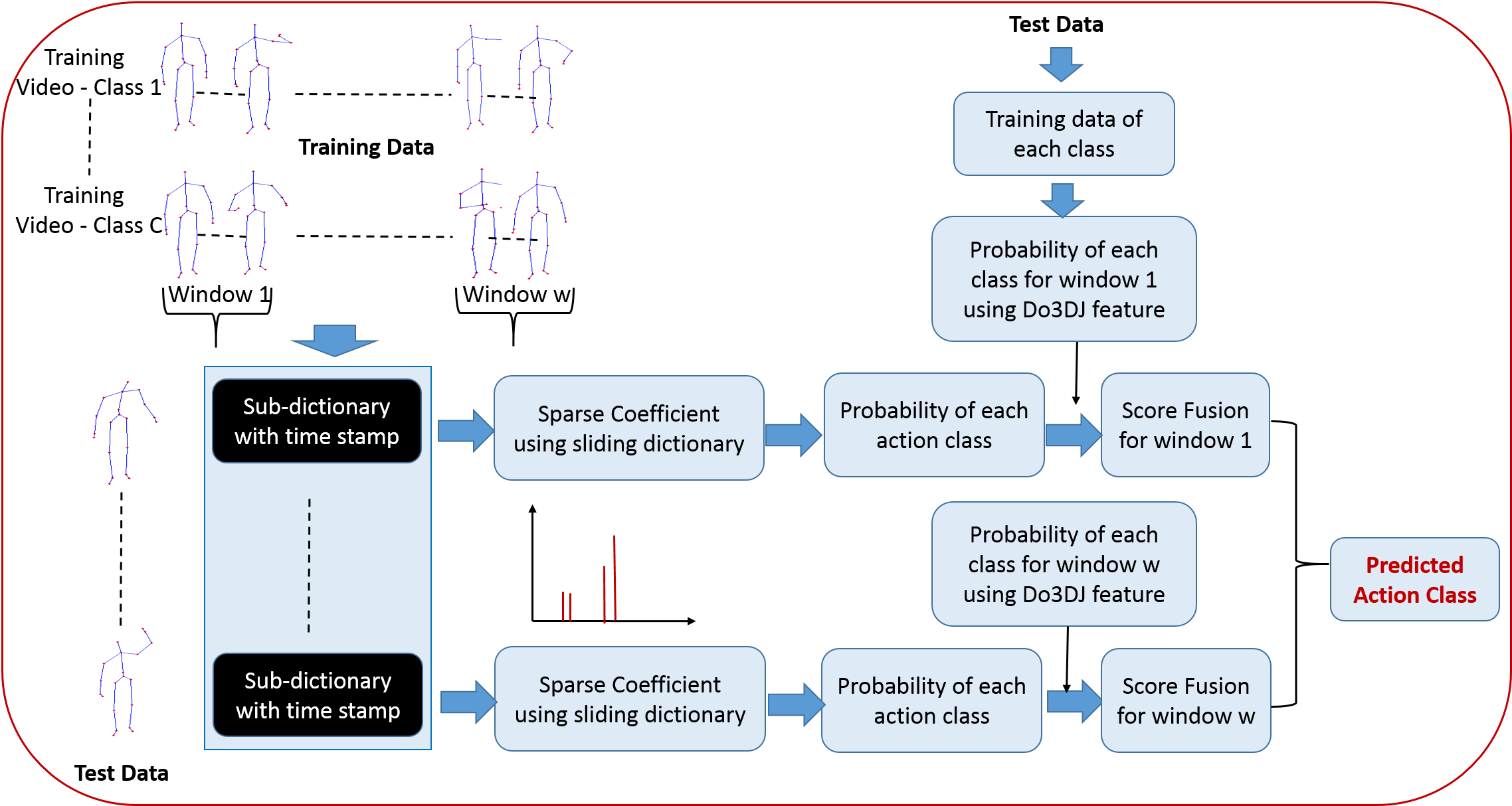}
	\caption{Flowchart of the proposed sliding window based sparse representation approach for online action recognition. Do3DJ denotes the difference of 3D joint features which is also used for computing the final confidence scores.}
\label{fc}
\end{figure*}

\section{Proposed Approach}
\label{prop}
Here, we describe the proposed approach for the task of action recognition.
First, we describe the sliding window based sparse representation framework, followed by the difference of 3D Joints feature and score fusion of the two features.

\subsection{Sliding Dictionary - Sparse Representation}
To make the proposed approach applicable for both offline and online action recognition, we propose a sliding dictionary-based sparse representation of the input action sequence.
Let $C$ be the number of different action classes and $K_c$, $c = 1,2,\hdots, C$ be the number of training sequences of class $c$. Let $W$ denotes the total number of overlapping windows for each action sequence.
The number of frames for each action sequence may be different and depends on the total number of frames in that sequence.
This is for offline action recognition when we know the total number of frames apriori.
Let $f^{c}_w$ denote the collection of all the features vectors for class $c$ and window $w$ given as $f^{c}_w = \{f_{w}^{c,1}; f_{w}^{c,2}; \hdots; f_{w}^{c,K_c}\}$. 
Here $f_{w}^{c,i}$ denotes the feature vector of the $i^{th}$ training example of class $c$ for window $w$.
Thus the combined data for all the classes for window $w$ will be denoted by 
\begin{equation}
f^{1, \hdots, C}_w = \{f_{w}^1; f_{w}^2; \hdots; f_{w}^C\}
\end{equation}
The complete dictionary is constructed from the training feature vectors of all classes and all the windows as follows:
\begin{equation}
 D =  [f^{1, \hdots, C}_1; \hdots; f^{1, \hdots, C}_w; \hdots; f^{1, \hdots, C}_W]   
\end{equation}
So the dictionary consists of all the features computed from all the training action sequences which are time stamped using their window index.

Given a test action sequence, it is similarly divided into overlapping windows and the feature vector corresponding to each of the windows $f_w^t$ is computed.  
For computing the corresponding sparse coefficients, instead of using the whole dictionary, we use a sliding dictionary based on the window index of the input sequence.
For example, for computing the sparse coefficient for $f_w^t$, the dictionary used is as follows
\begin{equation}
D_w=[f^{1, \hdots, C}_{w-N}; \hdots; f^{1, \hdots, C}_{w}; \hdots; f^{1, \hdots, C}_{w+N}] 
\end{equation}
i.e. the dictionary elements corresponding to window $w$ and $N$ windows before and after $w$ are used.
This sliding window ensures that the temporal evolution of the sequence is maintained, i.e. the initial part of the test sequence is not matched with the last part of a training sequence.
Multiple windows of the training sequence is considered in the dictionary to handle the temporal misalignments and rate variations in the action sequences. 
The corresponding sparse coefficient $\boldsymbol{\alpha_w}$ is obtained by solving the following
    \begin{equation}
    \argmin_{\alpha_w} || f_w^t - D_w \alpha_w||^2_2+\lambda || \alpha_w||_1
    \end{equation}
Here, we use the standard sparse coding solver SPAMS~\cite{spams} to solve for the sparse coefficient $\alpha_w$.

For the window $w$, we compute the probability of the test window belonging to each of the action classes using the reconstruction error.
Let $D_{w,c}$ be the matrix obtained from $D_w$ corresponding to the dictionary atoms belonging to action $c$ and $\alpha_{w,c}$ be the corresponding sparse coefficient obtained from $\alpha_{w}$ corresponding to the same action $c$. 
The reconstruction of $f_w^t$ using dictionary atoms corresponding to only class $c$ is given by: 
    	$\hat{f}_{w,c}^t=D_{w,c} \alpha_{w,c}$
   	 The reconstruction error for window $w$ for the action class is computed by taking the euclidean distance of the reconstructed feature and the original feature as
   	 \begin{equation}
   	 R_{w,c}=||	\hat{f}_{w,c}^t - f_w^t||^2
   	 \end{equation}
Lower reconstruction error for class $c$ implies that it is more likely to belong to that class as compared to classes for which the reconstruction error is high.
The probability $P^{dict}_w(c)$ that window $w$ of the test sequence belongs to the action class $c$ using the sliding dictionary is thus given by
\begin{equation}
P^{dict}_w(c)=\dfrac{\dfrac{1}{R_{w,c}}}{\sum_{k=1}^{C}\dfrac{1}{R_{w,k}}}
\end{equation}
The reason behind computing the probability of all classes instead of assigning it to a particular class is that a small part of a sequence  may appear similar to many action classes. For example, if we look at just the initial part of a walking and jogging sequence, they may appear very similar, which means that only a small segment is not sufficient to infer the class label.
But as we see more and more windows, the probability of the action belonging to the correct class will increase and that of the incorrect classes will decrease. \\ \\
{\bf Feature Used:} Since the framework presented above is general, any appropriate feature can be used as the input.
In this work, we have used covariance descriptor of the 3D skeletal joints as the input feature.
Let ${ n_{ w}}$ be the number of frames in ${ w^{th}}$ window $\{ w=1,2,3,...,W\}$ and ${J}$ be the number of joints. 
Let ${{ S{^i}}}$ be a ${J}\times 3$ dimensional matrix such that each row is a vector of $x, y, z$ co-ordinates of all the ${J}$ joints of  ${i^{th}}$ frame respectively. 
The covariance of the ${ w^{th}}$ window is calculated as:\\
   \begin{equation}
  Cov_w=\dfrac{1}{n_w}\sum_{i=1}^{n_w}(S^i-\mu^i_m)(S^i-\mu^i_m)^T
   \end{equation}
   	 where $\mu^i_m$ is the sample mean of $S^i$  and $(.)^T$ is the transpose operator.
   	  ${Cov_{w}}$ is a symmetric matrix by definition. Hence, only the upper triangular matrix is taken and concatenated to get a single feature vector $f_{w}$ of dimension $J+1\choose 2$ which is the covariance descriptor~\cite{hussein2013human}. 
$f_{w}$ is finally normalized to have a unit norm. 

 \subsection{Difference of 3D Joints Feature}
In this work, we augment the sparse representation based feature with another very simple feature, which is the difference of 3D joints.
We show that the combination of these two simple features is very effective for the task of action recognition and compares well with the state-of-the-art for several available datasets.
For this feature also, given an input video sequence, we divide it into overlapping segments as done for the previous feature.
This is done for both the training sequences as well as the testing ones.
For this feature, the probability of a test window belonging to a given action class is computed from the difference of the joint locations of the test window with all the training sequences of that class. 
For a test sequence, to compute its distance from a particular training sequence, we first compute the baseline difference which essentially aligns the two sequences in terms of their joint locations in the first frame.
Let $\phi_{t,0}$ and $\phi_{{k,0}}$ refers to first frame of testing and training respectively, i.e the neutral pose position, then this baseline difference is given by
\begin{equation}
\beta_{k}=\phi_{t,0}-\phi_{{k,0}}
\end{equation}
Now, for the $w^{th}$ window of the test sequence, we consider the $W_1 = w-N, \hdots, w+N$ windows of the training sequences.
We compute the differences between the joint locations of the frames in the test sequence window with all the frames in the selected window of the training sequence.
The distance between $n^{th}$ test frame of window $w$ from a frame in the chosen window of the training sequence is computed as (where $\phi_{k,n_1}$ is the joint locations of a frame of the training sequence in the chosen window)
\begin{equation}
\gamma_{k,w} = \phi_{t,n} - \phi_{k,n_1} - \beta_{k}
\end{equation}
For each action class, the $L$ least distances from the training sequences of that class $c$ are accumulated and the mean of these is computed which is denoted by its score $S_{w,c}$.
This is used to further compute the probability of the window $w$ of the test sequence of belonging to class $c$, i.e.
\begin{equation}
P^{3Ddiff}_w(c)=\dfrac{\dfrac{1}{S_{w,c}}}{\sum_{k=1}^{C}\dfrac{1}{S_{w,k}}}
\end{equation}


\subsection{Score Level Fusion}
 We perform score level fusion by combining the probabilities of both the sparse representation feature and difference of 3D joints feature. 
Then the final score for window $w$ is obtained as follows
\begin{equation}
   \tau_w(c)=\mu_1 P_w^{dict}(c)+\mu_2 P^{3Ddiff}_w(c) 
   \end{equation}
  \begin{center}
  	such that $ \mu_1+\mu_2=1$
  \end{center}
  $\mu_1$ and $\mu_2$ are parameters which are used to obtain trade-offs between the two scores. 
This is the confidence score that window $w$ of the test sequence belongs to action class $c$.
Suppose, $W_1$ windows of the test sequence is available (for online recognition with partial data, $W_1$ may be less than the total number of windows for the entire test sequence).
Then the final confidence score for all the action classes is computed and the predicted action class is given as follows:
\begin{equation}
\arg \max_{c} \prod_{w=1}^{W_1} \exp(\tau_w(c))
\end{equation}
   
   
\section{Experimental Evaluation}
\label{expts}
Now, we perform extensive evaluation of the proposed approach on $3$ publicly available datasets :\textit{UTD-MHAD}, \textit{ UT Kinect Action dataset} and \textit{ MSRC-12 Gesture Dataset}.

\subsection{ UT Kinect Action Dataset}
\begin{figure}[h]
	\centering
	\includegraphics[width=0.9\linewidth]{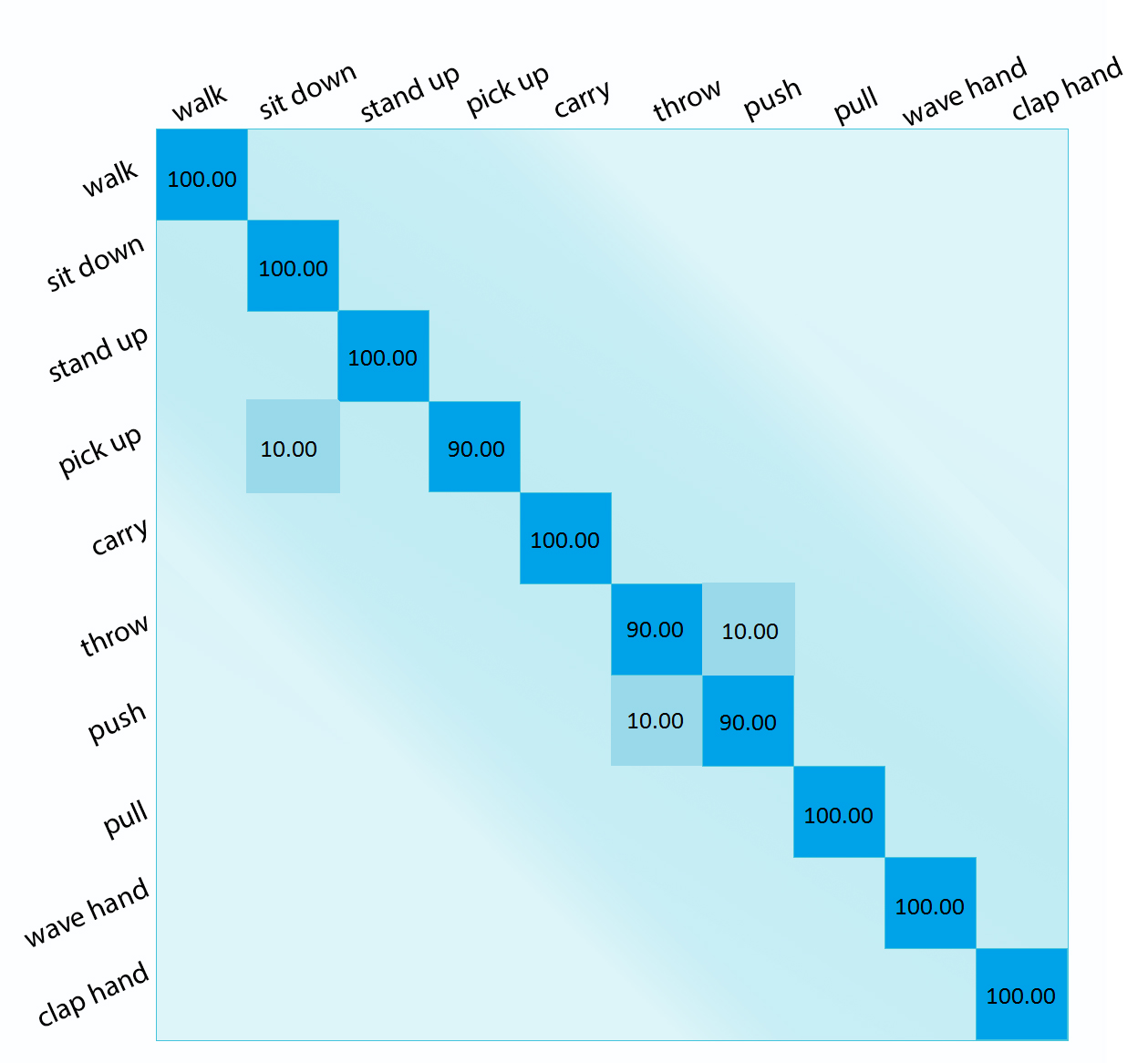}
	\caption{Confusion matrix for the UT Kinect Action Dataset.}
\label{confmat}
\end{figure}

The UT Kinect dataset~\cite{xia2012view} consists of 10 subjects performing 10 actions and each subject performs every action two times. 
This is a challenging dataset due to high variations in the actions of same class. 
The subjects 1, 3, 5, 7, 9 are used for training and 2, 4, 6, 8, 10 were used for testing, similar to the cross subject set-up of~\cite{zhu2013fusing}. 
The confusion matrix for this dataset is shown in Figure~\ref{confmat}.
We see that except for three actions, the approach performs perfectly for all the other actions.
Table~\ref{utk_table} reports the results of the proposed approach and the other state-of-the-art approaches for action recognition.
The other results are directly taken from~\cite{vemulapalli2014human}.
We observe that the recognition accuracy achieved by the proposed approach is comparable with the state-of-the-art~\cite{vemulapalli2014human} and significantly higher than all the other recently proposed approaches.

\begin{table}[h]
	\centering
	\caption{Recognition Accuracy of UT Kinect Action Dataset (\%).}
	\label{my-label}
	\begin{tabular}{|c|c|}
		\hline
		Histograms of 3D joints~\cite{xia2012view}    & 90.92   \\ \hline
		Random forests~\cite{zhu2013fusing}                & 87.90   \\ \hline
		Lie Group~\cite{vemulapalli2014human}                 & 97.08   \\ \hline
		\bf{Proposed Approach}                 & {\bf 96.97}   \\ \hline
	\end{tabular}
\label{utk_table}
\end{table}


\subsection{UT Dallas Multimodal Dataset}
The UTD-MHAD dataset~\cite{utd} is a very new dataset and is gathered using both Microsoft Kinect sensor and a wearable inertial sensor. The dataset consists of $27$ actions performed by $8$ subjects ($4$ females and $4$ males). 
Each action is performed by each subject $4$ times. 
Out of the total of $864$ sequences, we remove the $3$ corrupt sequences as in~\cite{utd} and use only the the skeletal positions of the remaining  $861$ sequences.
It is a comparatively difficult dataset as large number of actions are pooled together. 

We employ the experimental protocol of~\cite{utd}, and half of the subjects (odd numbered) are used for training and rest half of the subjects (even numbered) for testing. 
Table~\ref{utd_table} shows the results obtained using the proposed approach and also comparisons with the state-of-the-art.
Note that the result obtained using collaborative representation classifier method is 79.1\% using both kinect and inertial data.  However by using only Kinect it was found be 66.1\%. 
Also, Local Binary pattern is performed on depth maps, which are less noisy as compared to skeletal data. 
We see that the proposed approach outperforms both the other approaches and gives the best result for this dataset. 
\begin{table}[h]
	\centering
	\caption{Results on the UTD - MHAD dataset (\%).}
	\label{my-label}
	\begin{tabular}{|l|l|}
		\hline
		Collaborative representation classifier~\cite{utd} & 79.1    \\ \hline
		DMM-LBP{~\cite{chen2015action}}                                 & 84.2    \\ \hline
		\bf	{Proposed Approach}                              & \bfseries{86.12} \\ \hline
	\end{tabular}
\label{utd_table}
\end{table}


\subsection{MSRC - 12 Gesture Dataset }
The MSRC-12 Kinect gesture dataset~\cite{msr12} contains sequences of human movements, represented as body-part locations, and the associated gesture is to be recognized by the system. 
The dataset consists of 594 sequences and a total of 6,244 gesture instances. 
The motion files contain tracks of 20 joints estimated using the Kinect Pose Estimation pipeline. 
The body poses are captured at a sample rate of 30Hz with an accuracy of about two centimeters in joint positions. 
This is a very large dataset and is a good test of scalability of the proposed approach. 

We have used the test set-up of~\cite{hussein2013human}, where half of the subjects were used for training and half of them for testing. 
The experiment was repeated for 20 times, each time taking half of the subjects at random and we report the average over all the 20 iterations. 
The results of the proposed approach and comparison with the approach in~\cite{hussein2013human} is shown in Table~\ref{msr12_table}.
We see that for this dataset also, the proposed approach performs better than the state-of-the-art result.
\begin{table}[H]
	\centering
	\caption{Recognition Accuracy of  MSRC-12 Kinect Gesture Dataset (\%).}
	\label{my-label}
	\begin{tabular}{|l|l|}
		\hline
		Covariance descriptors~\cite{hussein2013human}         & 91.7      \\ \hline
		\bfseries{Proposed Approach}                      & {\bf 92.89}        \\ \hline
	\end{tabular}
\label{msr12_table}
\end{table}

\begin{figure*}[t]
	\centering
	\includegraphics[width=0.9\linewidth]{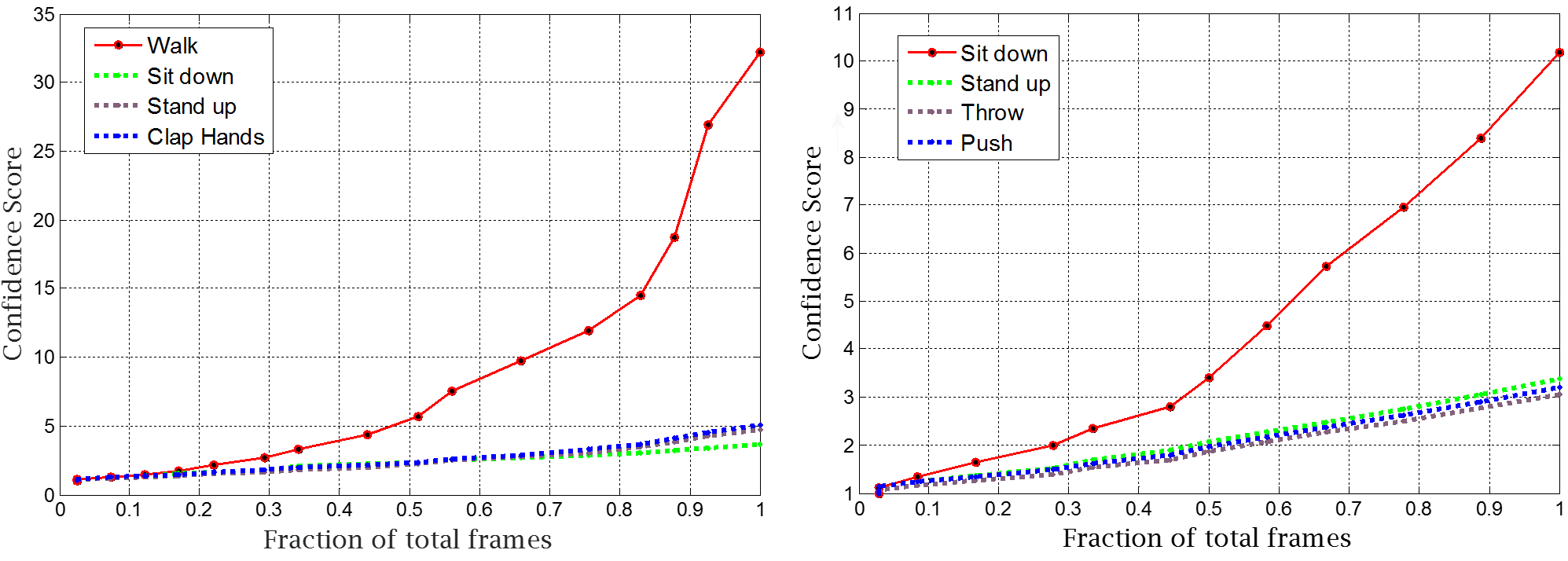}
	\caption{The scores for 4 different actions for UTK dataset. The correct class is walk (left) and sit down (right). The scores for the correct class increases and those of the wrong classes decrease as more and more frames become available.}
	\label{update_fig}
	\vspace{-0.2cm}
\end{figure*}
\subsection{Online Action Recognition}
	 The above experiments were performed using the whole action sequence and compared with the state-of-the-art in the offline scenario.
	Since the proposed approach relies on incremental update of the probability of each action class with the availability of each window, it adapts seamlessly to online action recognition as well. 
In this work, for both online and offline action recognition, we assume the action has already been detected from an unsegmented video using a suitable detection approach as in ~\cite{sharaf2015real},~\cite{zanfir2013moving},~\cite{msr12}.  This detection part would provide the action points on the unsegmented sequence and the beginning and ending point of an action. 
All further computations required for the recognition would be carried out by our algorithm, even with the availability of partial data. 

The main difference between online and offline version of action recognition is that for the online scenario, only a partial number of frames are available and the recognition has to be done using the incomplete information.
The total number of frames in the sequence is also not known apriori, unlike the offline scenario.
In this work, we evaluate the usefulness of the proposed algorithm for online action recognition on the offline datasets itself, \textit{UT Kinect} and \textit{UT Dallas}, under the constraint that only partial data is available.
Since the total number of frames is not known beforehand, we perform a frame-level computation of the probabilities. The training sequences are divided into overlapping windows as before, but for the testing sequence, we consider window sizes of variable length.
For each frame, we take the maximum probability among all the windows which has the particular frame as the middle one to be probability of that frame belonging to a particular action class.
The score level fusion is performed similar to the offline scenario.
As more and more frames are available, the probability confidence score keeps on accumulating, and the score increases for the correct class and decreases for incorrect classes, until the detection algorithm gives the 'end of action' signal. 
We achieve a recognition accuracy of 88.89\% on UTK Dataset and 79.07\% on UTD Dataset using the covariance descriptor as the feature to the dictionary. The online scenario using these datasets is challenging due to the fact that only partial data is available.  
We observe that the recognition accuracy decreases marginally from the offline to online scenario, thus proving the effectiveness of the seamless adaptability of the proposed approach.  

Since the proposed approach is based on incremental update of confidence scores, first we perform an experiment to see how the scores gets updated for some example actions of the UT Kinect Action dataset.
Figure~\ref{update_fig} shows how the scores are updated as a function of the number of sliding windows for two different actions, for test actions walk (left) and sit down (right) respectively.

Two noteworthy observations can be made from the above mentioned plots:
	\begin{enumerate}
		\item As more frames become available, the confidence (score) of the correct action class increases as compared to the incorrect action classes.
		\item Depending on the relative pace at which score of the correct class outgrows those of the incorrect classes, partial number of frames may be sufficient to get the correct class. 
	\end{enumerate}
	 We note from the experimental results that even with very few frames, the approach predicts with reasonable accuracy the correct action class, signifying its applicability for online action recognition task.
We report results on the UT Kinect Action dataset, but we have observed similar behaviour for the other datasets as well.
	\begin{figure}[t]
		\centering
		\includegraphics[width=\linewidth,height=5 cm]{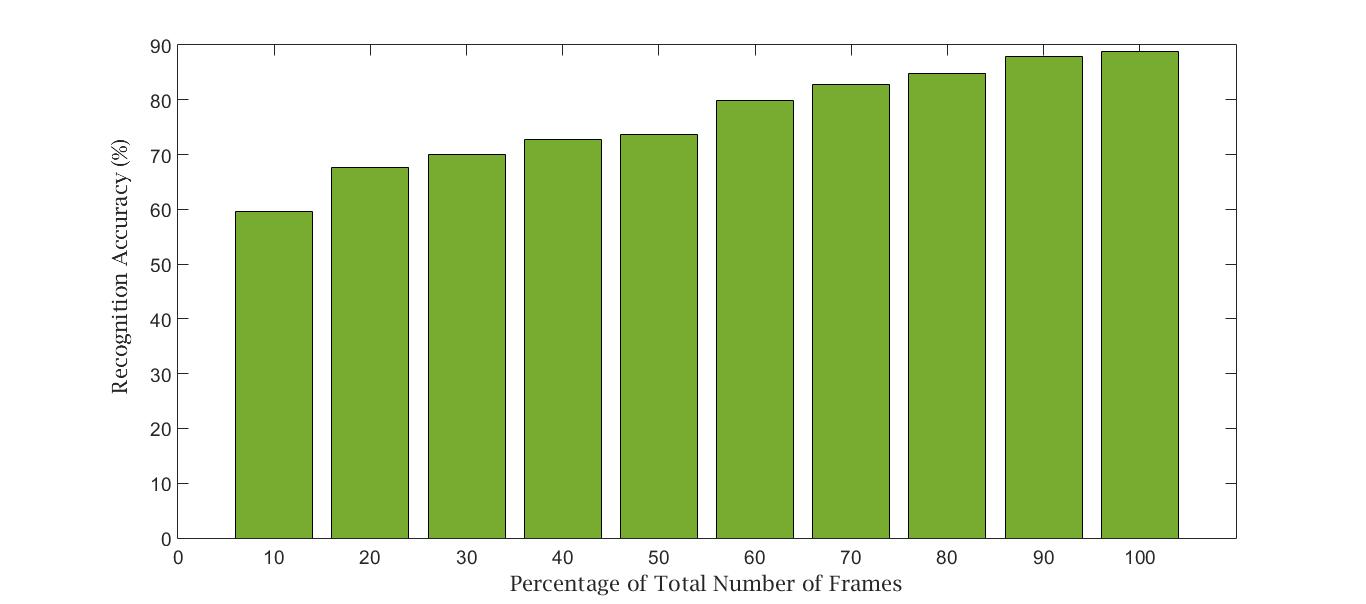}
		\caption{Recognition accuracy as a function of total number of frames for the UTK dataset.}
		\label{onlineutk}
	\end{figure}
		
	We perform another experiment to observe the number of frames that are required to get a reasonable accuracy for action classification. We consider increasingly more number of frames and plot the recognition accuracy.
	Figure~\ref{onlineutk} shows the recognition accuracies of the UTK dataset as a function of the total number of available frames in the whole video sequence.
	We observe that at around $70\%$ of the total number of frames, the recognition accuracy reaches close to the highest accuracy obtained by the proposed algorithm. Also, we get a $59.6\%$ accuracy with only $10\%$ of frames availability.
	We observed similar performance for  the other datasets also.
	This justifies the usefulness of the proposed approach for online action recognition.

\section{Conclusion and Future Work}
In this paper, we have presented an algorithm which employs covariance of 3D Joints descriptor to construct a sliding dictionary. This dictionary is designed such that temporal variations are accounted for. We have introduced a frame level difference of skeletal joints, which calibrates the testing action.
Score level fusion of the two scores gives the final confidence score for each action class.
Extensive experiments on different datasets justify that despite being simple, the proposed approach is very effective for both online and offline action recognition.

The proposed framework is general and more suitable features can be seamlessly added to improve the recognition accuracy of the proposed approach and this will be one direction of our future work.
Also, in future, we would like to use dynamic time warping to better handle the temporal misalignments, which can result in further boost in the accuracy.
Another direction of future work is to extend this approach for actions with lateral shifts, for e.g. same action done with the right hand once and left hand the other time. 
We would like to generate synthetic training data to take care of this problem as part of our future work.


{\small
\bibliographystyle{ieee}
\bibliography{papercite}
}

\end{document}